\documentclass[10pt,a4paper]{article}
\usepackage[utf8]{inputenc}
\usepackage{amsmath}
\usepackage{amsfonts}
\usepackage{amssymb}
\usepackage[dvips]{graphicx}
\usepackage{here}
\usepackage[margin=2.5cm]{geometry}
\author{Jean SALLANTIN\\ \texttt{js@lirmm.fr}  \and Antoine SEILLES\\ \texttt{Antoine.Seilles@lirmm.fr}  \\\texttt{LIRMM-CNRS UMR 5506 161 rue Ada 34392 Montpellier Cedex 5 - France}}
\title{n-opposition theory to structure debates}
\begin{document}

\maketitle
\begin{abstract}
\indent 2007 was the first international congress on the “square of oppositions”. A first attempt to structure debate using n-opposition theory was presented along with the results of a first experiment on the web. 
Our proposal for this paper is to define relations between arguments through a structure of opposition (square of oppositions is one structure of opposition). We will be trying to answer the following questions: How to organize debates on the web 2.0? How to structure them in a logical way? What is the role of n-opposition theory, in this context?
\\
We present in this paper results of three experiments (Betapolitique 2007, ECAP 2008, Intermed 2008).
\end{abstract}

\section*{Introduction}
\indent In the era of participatory democracy and web2.0, does a solution to debate exist on the web? Is it even possible? Existing solutions are very limited (chats, forums...). Many sites exist, for Web-users to express their opinion, but can they build structured arguments and confront them? We try to answer these questions by offering new software based on logic. But then, what kind of logic can describe the arguments of a debate?
What is the role of n-opposition theory?
\\
\indent In this paper we compare the results of three experiments. First we will introduce some important notions, then
we will present our experimental protocol and finally we will present detailed discussion system. We will also explain how we use the n-opposition theory to structure debate and reason. To conclude, we will discuss experiments contexts and results.

\section{Toward a web2.0 debate}
\indent The work  we will present began in 2007. Today it is related to the Intermed project. The Intermed project aims to provide a set of interoperable tools to debate on the web. This first chapter presents our actual model of debate and define some important notions.

\subsection{Modeling}
\indent We did not create any debate model for our first experiment (Betapolitique 2007). All we had was a set of documents (political articles of betapolitique website www.betapolitique.fr) and users had the opportunity to annotate and express their point of view. We present Annotation in the next part of this chapter. The users point of view or argument as we will define it later, comprised part of a text, a judgment and a comment.
\\
\indent For our second experiment (ECAP 2008 www.lirmm.fr/ECAP08), the documents are the texts of the congress speakers. Arguments are more structured and selected text can now be rephrased.
\\
\indent For our third experiment (the first Intermed experiment), we had the opportunity to work with
N.Desquinabo (phD in Cemagref UMR G-Eau ). N.Desquinabo thesis is about interlocutivs kinds in TV shows
\cite{NicoDeskiThese}. N.Desquinabo tasks in the Intermed project revolve around interlocutivs kinds in debates on
web \cite{NicoDeskiPaper}. This collaboration lead us to produce a first relational model based on the
following principles :

\begin{itemize}
 \item A \textbf{debate} is firstly defined by a subject and reflects a problem (to be resolved or restate …).
 \item Each debate is associated with a \textbf{body of documents} called references.
 \item Documents can be put in \textbf{categories}.
 \item Participants can create \textbf{group} to defend a same opinion.
 \item Participants can link \textbf{arguments} to a document.
 \item \textbf{Relations} link arguments together.
\end{itemize}

Some ideas in \textit{le discours acteur du monde} by G.Vignaux \cite{Ldam} illusrates our approach.

\begin{itemize}
 \item p55 : An argumentation is a set of reasoning to enhance a thesis. It means argumentations are done to solve, enhance or defend and an argumentation is firstly defined by a problem.
 \item definition of the dialectical discussion p30 quoted by G.Vignaux from the introduction of the Aristotle's Topiques by Brunschwig \cite{bruns}: A two players game; It's a binary opposition, affirmation and negation, victory and defeat.
 \item p57 : Argumentation consists in confronting arguments and find an issue.
\end{itemize}

\subsection{Arguments and annotation}
\indent First we present our mecanism to write arguments based on annotation. An argument is a statement supporting a
thesis or against a thesis. A thesis is a point-of-view regarding a debate. A thesis can be, for example, an opinion
(for or against Nanotechnologies?) or an important conclusion (Nanotechnologies are dangerous) that is to
be argued. Arguments are judgments that are expressed regarding a statement. A set of arguments for a thesis is called
an argumentation.
\\
\indent Today forums, chats or emails are used to debate via the web. We have decided to build a different system
using annotation because it can precisely localize the part of a text that a user wants to comment (or argue upon).

\begin{verse}
 Definition of \textbf{Annotation} : An annotation is a data attached to a document or a part of a document.
\end{verse} 

\indent We had first developped ad-hoc annotations systems for our first and second experiment platforms
(Betapolitique was built with SPIP www.spip.net, and ECAP website was built with XWiki www.xwiki.com).
For our third experiment, we have developed a solution based on Annotea protocol (www.w3.org/2001/
Annotea) and Annozilla (Annozilla is a client for the Annotea protocol developped by Mozilla for firefox web
browser annozilla.mozdev.org). This solution is currently a prototype and is independent of website platforms.
\\
\indent There are other annotation systems existing (we recommend the lecture of \cite{toward} ) but choosing Annotea was done for the following reasons :
\begin{itemize}
 \item It allows to annotate HTML documents.
 \item Annotea is a w3c web standard protocol. This guarantees interoperability.
 \item Annotations can be stored in RDF database.
 \item This is an asynchronous solution based on client-server architecture.
\end{itemize}

\indent We have added to the current Annozilla implementation the possibility to design forms for annotation on a specified category of texts.
\indent In our experiments, an argument is an "illocutionary-act". The statement is selected in a document and must be rephrased to be judged, and user has to justify his act by a comment.

\subsection{Experimental protocol}
\indent	This section deals with the experimental protocol we have built from our experiments. Judgments and relations between judgments are defined with respect to the community of users, the subject of the debate, and the categories of documents. A body of documents is created. Users read documents and can annotate to express their point of view and argue. They select and rephrase a statement in a document, state their judgement on it and justify it. Then, each argument can be discussed. When the protocol ends, a summary of the debate must be produced.
\begin{enumerate}
 \item Define a subject for the debate.
 \item Create forms for annotation (define judgments users can choose).
 \item Reading documents.
 \item Production of arguments.
 \item Discussion of arguments (there is a loop between the fourth and fifth step cause this discussion leads to produce new arguments).
 \item End of the experiment, production of a summary
\end{enumerate}
\indent In a following section of this document, we present three experiments. Criterias given to compare these experiments are :
\begin{itemize}
 \item Subject of the debate.
 \item Categories of documents.
 \item Do users know each other ?
 \item Links between judgments.
 \item Aim of the debate (Is this debate going to produce a new document?).
 \item Duration of the experiment.
 \item Content of justifications (is there a model of a justification for a kind of judgment?).
\end{itemize}

\indent Aim of each experiment:

\begin{enumerate}
 \item First experiment was done as a test to see if it is natural to use judgments to annotate.
 \item Second experiment was done as a test to see if users rephrase their selection and justify their judgment.
 \item Third experiment was done to test the structure that is itself based on the n-opposition theory.
\end{enumerate}

\section{Relations between arguments and n-opposition theory}
\indent The annotation forms we have designed force users to make a judgement. We propose to use structures of the n-opposition theory to define relations between judgments. Experiments are to confirm if users naturally respect or not these structures (Do users respect the structure without explanation?).
\\
\indent This section presents the n-opposition theory and structures we designed for our three experiments.

\subsection{The n-opposition theory}
\indent To define relations between judgments, we design a n-opposition structure based on the n-opposition theory \cite{Gfm}. This theory is a geometrical generalisation of Aristotle's classical opposition theory. The most famous structure of Aristotle's opposition theory is Aristotle's logical square of oppositions.

\begin{figure}[H]
\begin{minipage}[b]{0.40\linewidth}
      \centering \includegraphics[scale=0.4]{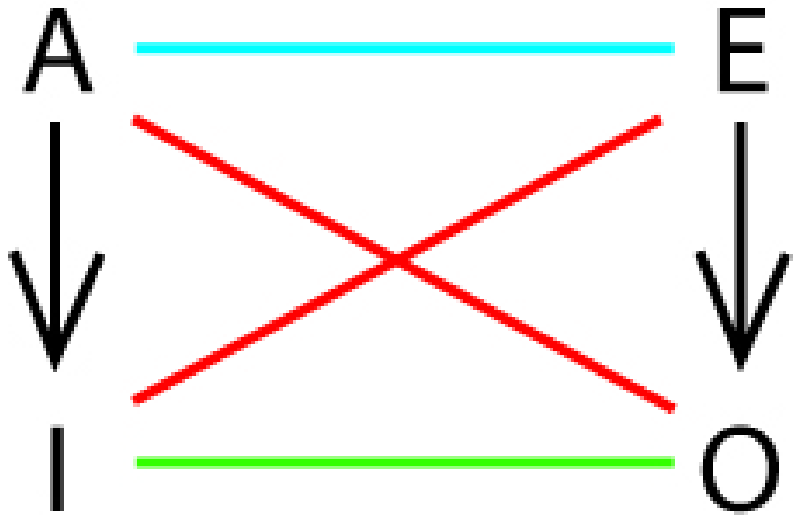}
      \caption{\it traditional Aristotle's logical square of opposition}
   \end{minipage}\hfill
\begin{minipage}[b]{0.40\linewidth}
      \centering \includegraphics[scale=0.4]{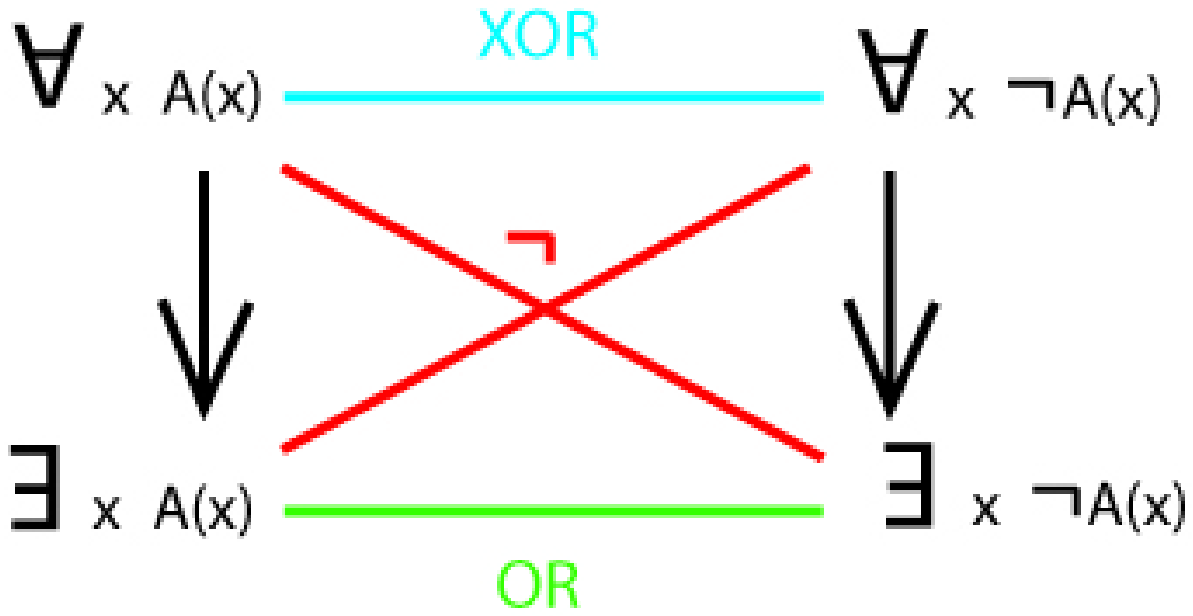}
      \caption{\it Aristotle's square of opposition in predicate logic}
   \end{minipage}\hfill
\end{figure}

Figure 1 shows four kinds of relations (arrows and colored line) between four judgments, traditionally represented by letters A,E,I,O. A and E represent universal judgments (A is the affirmative universal judgment, E is the negative universal judgment). I and O represent particular judgments (I is the affirmative particular judgment, O is the negative particular judgment). Arrows between universal and particular judgments represent relations of \textbf{sub-alternation}. The blue line between A and E represents a relation of \textbf{contrariety}. The green line between I and O represents a relation of \textbf{subcontrariety}. Red lines represent \textbf{contradiction}. Figure 2 shows an example of square of oppositions in predicate logic.
\\
\indent Following definitions are quoted from \cite{Las}.
\begin{itemize}
 \item \textbf{sub-alternation} : it is defined as the impossibility of having the first term true without having also the second true.
 \item \textbf{contrariety} : Contrariety between two terms is the impossibility that both terms are true, but the possibility that both terms are false.
 \item \textbf{sub-contrariety} : Sub-contrariety is the impossibility that both terms are false, but the possibility that both are true.
 \item \textbf{contradiction} : Contradiction between two terms is defined as the impossibility for them to be both true and both false.
\end{itemize}

\indent Today, our system doesn't assist users because we test if users naturally respect the structure. If a structure appears to be adapted for a debate and its users, it will become possible to assist users by signaling them when they violate the structure.
\\
\indent The following section presents structures we tested in our experiments.

\subsection{Structures of opposition}
\indent We defined six judgments for our first experiment. Two for a \textbf{theoretical} level : \textbf{i agree}, \textbf{i don't agree}. Two for a \textbf{practical} level : \textbf{it works}, \textbf{it doesn't work}. Two for an \textbf{emotional} level : \textbf{I like}, \textbf{i don't like}. We suggest a structure in which emotional level is a consequence of theoretical level and practical level is a consequence of emotional level. We considered instinct can be educated by theory and acts are consequences of instinct.

\begin{figure}[H]
\begin{center}
 \includegraphics[scale=0.5]{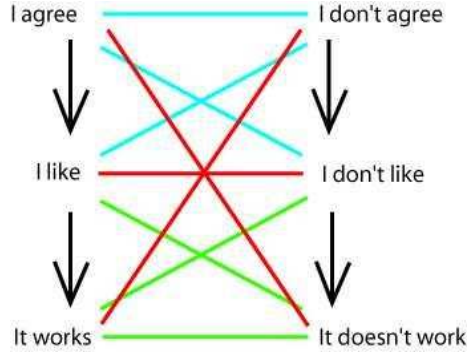}
\caption{\it structure of opposition for the betapolitique experiment}
\end{center}
\end{figure}

\indent D.Luzeaux \cite{Las} shows how to translate a n-opposition structure into an epistemistic logic. We had found modalitites of a defeasible logic introduced in \cite{Ffdl} organized in a n-opposition structure according to their definitions. This structure looks like the one we designed for the experiment. Defeasible reasonings deals with conflictual statements. Defeasible logics were introduced by D.Nute \cite{Dl}.
\\
\indent A defeasible theory is a triple (F,R,$>$) where F is a set of facts, R a set of rules and $>$ a superiority relation. There
are three kinds of rules, strict rules, defeasible rules and defeaters.

\begin{enumerate}
 \item Facts are a set of literals and represent indisputable knowledge .
 \item Strict rules are denoted by $A->B$ and are interpreted in the classical sense.
 \item Defeasible rules are rules that can be defeated by contrary evidences.
 \item Defeaters are rules that cannot be used to draw any conclusions. Their only use is to prevent 
conclusions. In other words, they are used to defeat some defeasible rule by producing evidence to the
contrary.
 \item A superiority relation on R is an acyclic relation $>$ on R. $r1>r2$ expresses that r1 may override r2 if they
lead to conclusive decision that contradict each other.
\end{enumerate}

\indent A conclusion in a defeasible theory D is a tagged litteral with one of the following modality:
\begin{enumerate}
\item \textbf{$+\Delta q$} means that q is definitely provable in D. q is proved with only facts and strict rules.
\item \textbf{$-\Delta q$} means that q is not definetly provable in D. q is not proved with only facts and strict rules.
\item \textbf{$+\delta q$} means that q is defeasibly provable in D. A defeater to contradict q can exist. The
superiority relation leads to conclude q before $\neg q$.
\item \textbf{$-\delta q$}  means that q is not defeasibly provable in D. No set of rules lead to conclude q or for every
set of rules leading to conclude q, the superiority relation leads to conclude $\neg q$ before.
\item \textbf{$+\delta _{ap} q$} means q is provable with ambiguity in D. There is a set of rules that can lead to
conclude q. The superiority relation is not taken into account.
\item \textbf{$-\delta _{ap} q$} means that q is not provable even with ambiguity in D. There is no set of rules that
can lead to conclude q.
\end{enumerate}

\begin{figure}[H]
\begin{center}
 \includegraphics[scale=0.5]{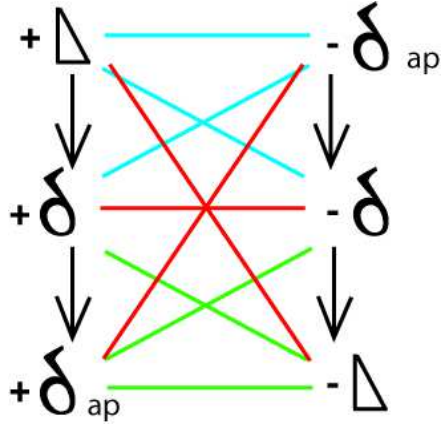}
\caption{\it structure of opposition for a defeasible logic}
\end{center}
\end{figure}

\indent However, we didn't look any deeper to confirm the matching between these two structures. This observation was the result of intuition. In our experiments, we focused on defining structures and testing their efficiency. Translation of n-opposition structures into logic is one of our future task.
\\
\indent For the second experiment (ECAP), we had defined six judgments:
\begin{itemize}
 \item \textbf{I agree}, \textbf{I don't agree} to judge a statement on its purpose.
 \item \textbf{Correct syntax}, \textbf{uncorrect syntax} to judge a statement on its syntax.
 \item \textbf{This confirms}, \textbf{this contradicts} to add supporting references or contradictory references to the statement.
\end{itemize}

\indent Judgments were just opposed by contradiction. The Aim of this experiment was to test if users would accept to rephrase their selections and justify their judgments and we didn't propose a complete structure.
\\
\indent For the third experiment (Intermed), we proposed three levels of judgments :
\begin{enumerate}
 \item Proved 
 \item Unclear
 \item incorrect
\end{enumerate}

\indent For each annotation, students had to judge with a combination of an affirmative or negative judgment per level (yes it is proved or no it is not proved and yes it is unclear or no it is not unclear and yes it is incorrect or no it is not incorrect). We tested relations of sub-alternation of the figure 5.

\begin{figure}[H]
\begin{center}
 \includegraphics[scale=0.5]{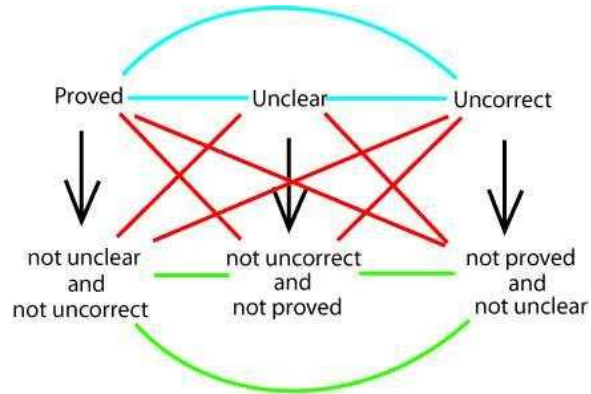}
\caption{\it structure of opposition for the Intermed experiment}
\end{center}
\end{figure}

\section{Experiments}
\subsection{Conditions of experiments}
\indent Before dealing with results and observations of experiments, we present context of these experiments. For the first experiment, we worked with the development team of betapolitique. Betapolitique is a web blog of politic news. The duration of experiment was three months (February to may 2007) during French presidential elections. No help nor explications were provided to users. There was no objective nor a subject of debate defined. Users of betapolitique are anonymous to each other and were not forced to participate. They could read their arguments on the right of the article (figure 6) and use a treemap (developed by the society Pikko www.pikkosoftware.com) to see which article was the most annotated (figure 7).

\begin{figure}[H]
\begin{center}
 \includegraphics[scale=0.4]{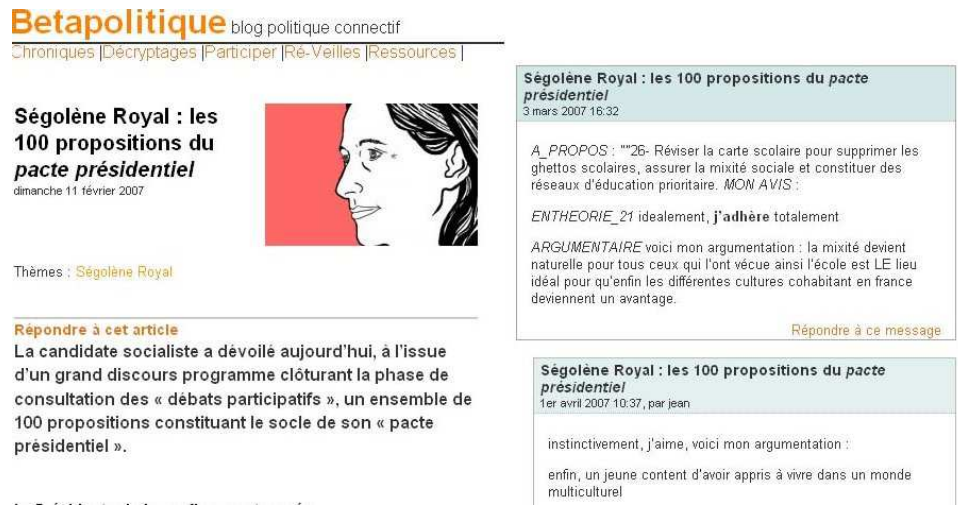}
\caption{\it screenshot of annotations for Betapolitique}
\end{center}
\end{figure}

\begin{figure}[H]
\begin{center}
 \includegraphics[scale=0.2]{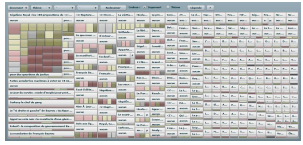}
\caption{\it A treemap to visualize annotated documents}
\end{center}
\end{figure}

\indent For the second experiment, we collaborated with the organization committee of ECAP congress. The web site of ECAP was designed to read and discuss texts of the speakers of the congress. The experiment begun few days before the congress and ended to the end of the congress (but the website still exists). A tutorial is accessible. Participants of the congress were unknown each other. Their arguments were displayed on the right of the text and corresponding selections were highlighted into the text (figure 8). The goal of the debate was to enhance the quality of presentations by taking into account remarks from others.

\begin{figure}[H]
\begin{center}
 \includegraphics[scale=0.7]{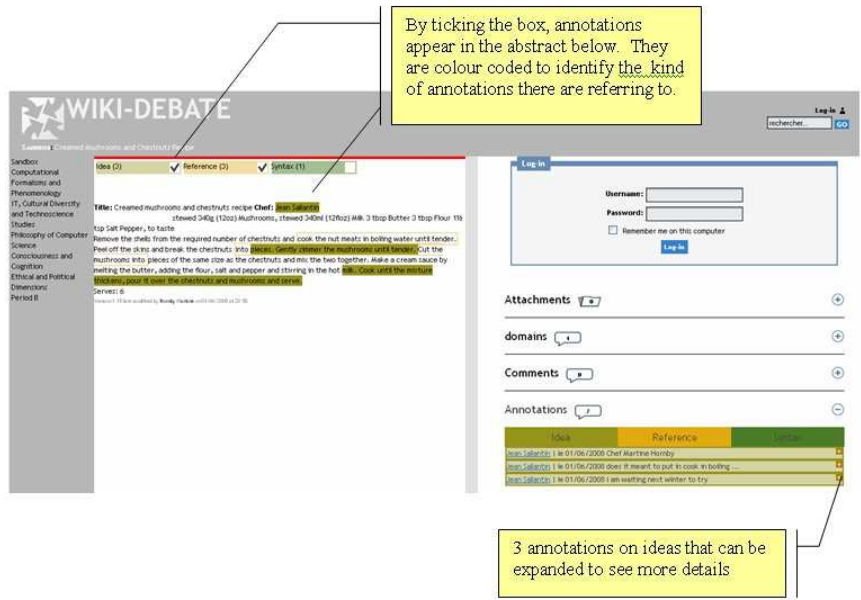}
\caption{\it screenshot of annotations for ECAP}
\end{center}
\end{figure}

\begin{figure}[H]
\begin{center}
 \includegraphics[scale=0.4]{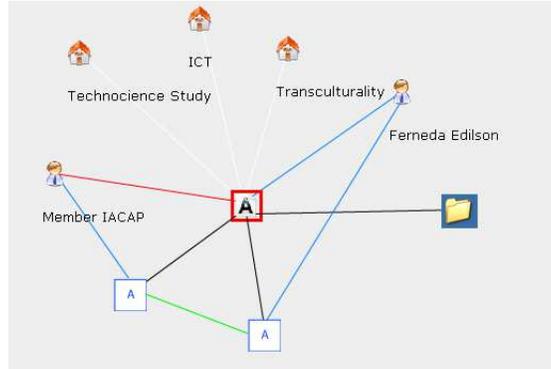}
\caption{\it A graph to cartography debates based on opposition}
\end{center}
\end{figure}

\indent For the third experiment, a group of experts of the domain of nanotechnologies wrote a body of documents to answer the following questions :
\begin{enumerate}
 \item Do nanotechnologies really bring innovations or is it just a buzzword for well known technologies?
\item Are nanotechnologies a chance for the developping countries, or is it still another divide between north and
south?
\item Is it sacrilege or progress to let nanotechnologies alter human nature?
\item Do the nanosciences and nanotechnologies harbor still unknown risks? What is a responsible attitude :
should research be stopped and the precautionary principle be applied, or should we be confident that
further research will identify the actual risks and provide means to control them?
\item Is the lack of proof that nanotechnologies are harmful or harmless a justification for pursuing production
and research, or is it a justification for stopping them?
\end{enumerate}

\indent These texts were the documents that students have annotated. The subject of the debate was : Should we write a moratorium on nanosciences and nanotechnologies? And the goal was to produce a summary of argumentations.
The experiment was done during two class sessions. Our team was present to assist them. Their annotations were listed on the left of the texts. The parts of text discussed were highlighted by a background color (fig 10).

\begin{figure}[H]
\begin{center}
 \includegraphics[scale=0.3]{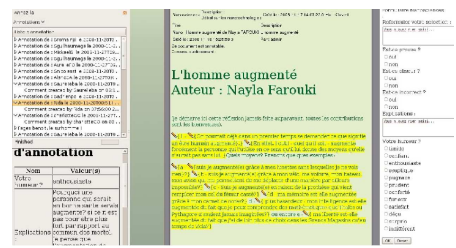}
\caption{\it A screenshot of annotations for Intermed}
\end{center}
\end{figure}

\begin{figure}[H]
\begin{center}
 \includegraphics[scale=0.4]{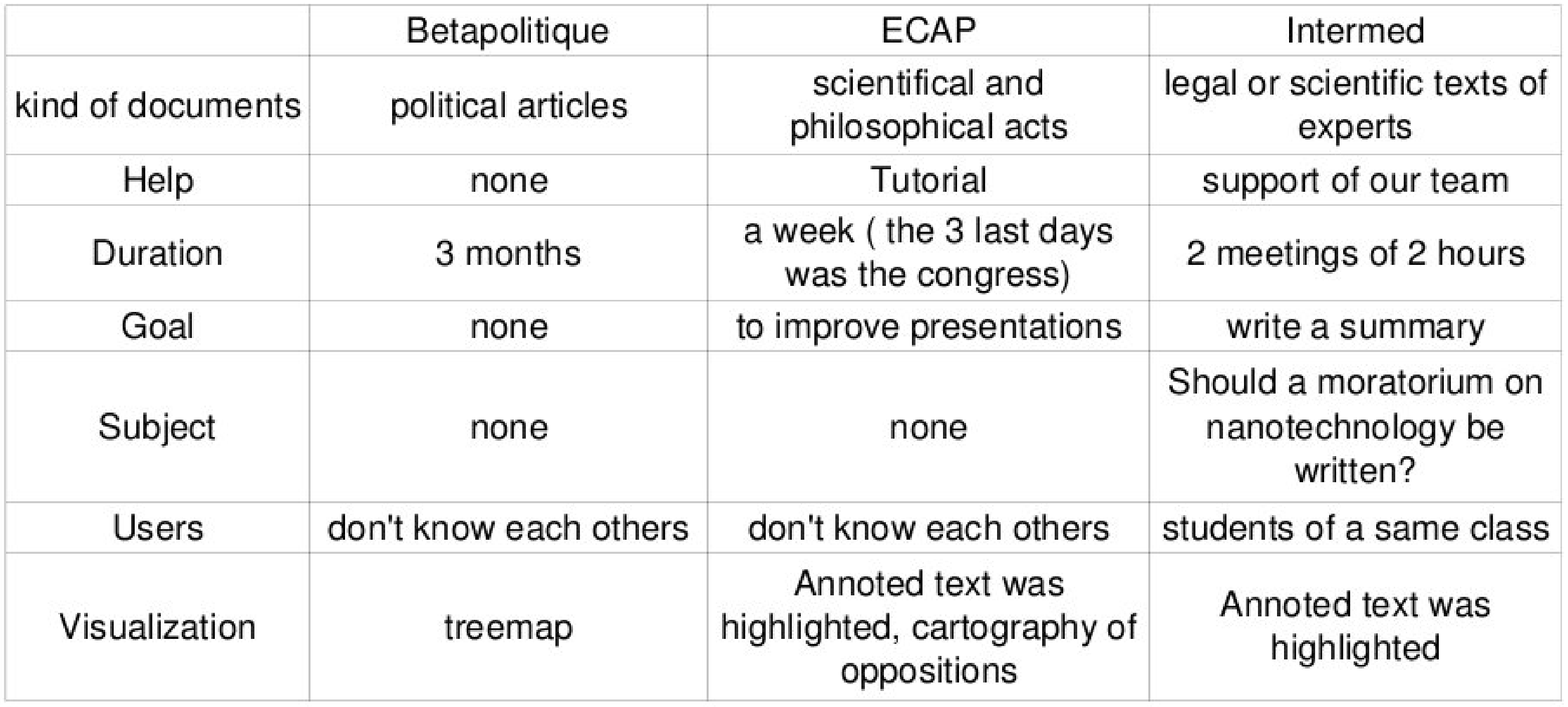}
\caption{\it Comparison of conditions of experiments}
\end{center}
\end{figure}

\indent Each of these experiments made evolve our experimental protocol. We decided to define a subject of debate for the third experiment. Maybe, to not define clearly a subject for the two first experiments was the reason of the lack of discussions whereas in the third experiment, each argument was discussed.
\\
\indent It appeared clearly for the second experiment that judgments have to be defined depending the community. For the third experiment, we defined judgments depending the subject of the debate. We are now convinced that judgments must be defined depending the categories of documents. People don't judge scientific statements the same way as philosophical statements.
\\
\indent With the second experiment we introduced the obligation to rephrase the selection. This enhances comprehension between users. A judgment is associated to the new statement (rephrasing the selection) , not to the selected part of document. So, even if there is a misunderstanding of the selected part of document, user precise the statement he judge.

\subsection{Observations and results}
\indent We present in this section the conclusions of these three experiments.
\\
\indent The first experiment confirmed the use of judgments and the choice of annotation as a solution for implementing debate.
We had 870 annotations for three months of experiment. Judgments were all adopted (figure 12).

\begin{figure}[H]
\begin{center}
 \includegraphics[scale=0.4]{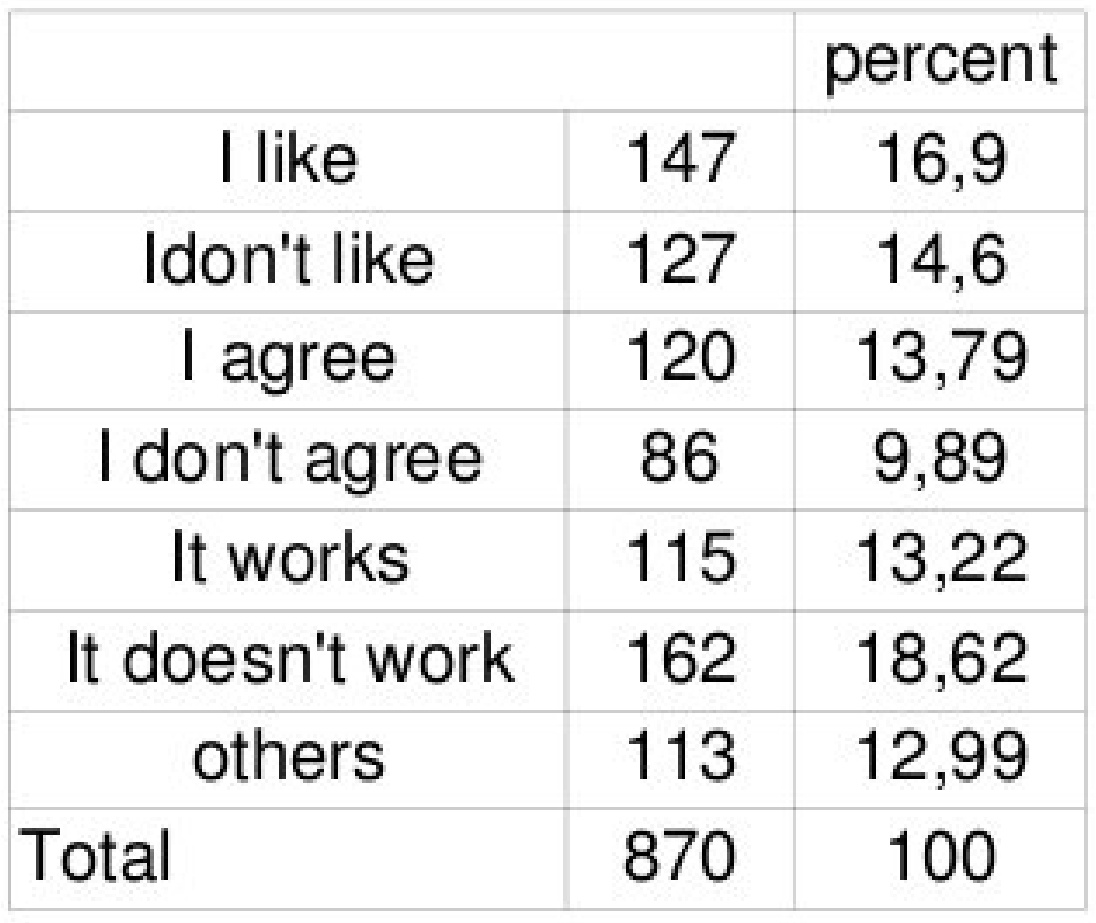}
\caption{\it Repartition of judgments in betapolitique}
\end{center}
\end{figure}

\indent The second experiment validated the need of an ergonomic interface and of a gratification function. The goal of the experiment was not taken into account by participants. We did interviews as well as polls and the general feeling was that the idea is interesting but the tool is too complicated to be used.
We analysed frequencies of visit on the site through Google analytics tool (www.google.com/analytics). The Congres was held from the 16th to the 18th of June 2008. The highest number of visits appened on the 16th of June, 147 visitors for 80 participants registered to the congress. We had 306 visitors from the 15th to the 19th of June. The average time of visit during this period was of four minutes. Though we had so much visitors, we had only 95 annotations, most of them done before 19th June. The text the most annotated has been annotated 7 times (there were 70 texts). Nearly nobody attempted to discuss through the tool.

\begin{figure}[H]
\begin{center}
 \includegraphics[scale=0.35]{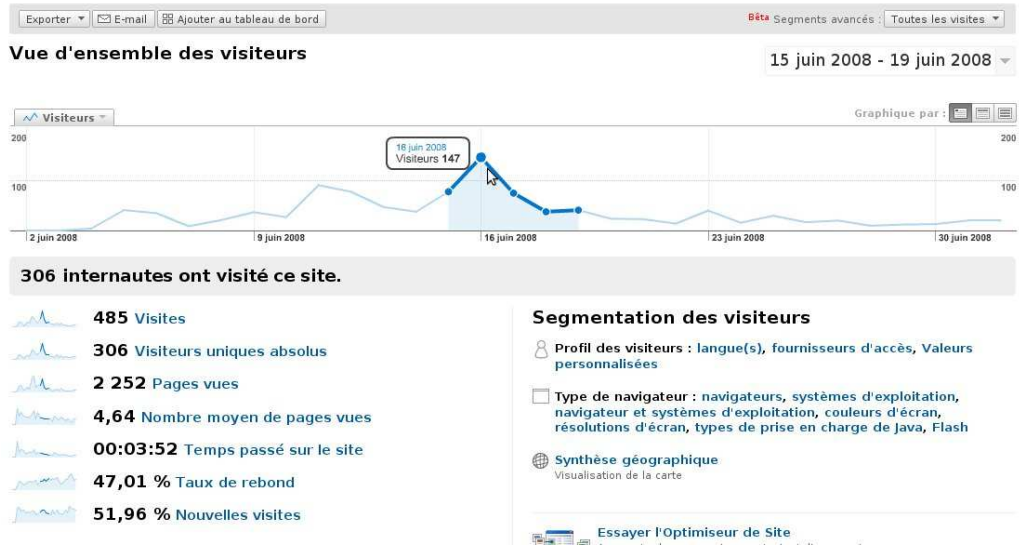}
\caption{\it Statistics of visits of ECAP website}
\end{center}
\end{figure}

\indent The third experiment validated the interest of rephrasing the annotated statement. The most interesting observation is the fact that there seems to be models of justifications for judgments.
Users adopt the same kind of locution for a same combination of judgments to justify their choice. Satisfaction, interest and evolution of the opinions were evaluated with a survey. Most of students have appreciated the quality of debates using our tool. They suggested enhancements for the tool that are possible with our technological and conceptual choices.
\\
\indent Most of the arguments are of ethical content. They are on managing risks or on managing conflicts between nanotechnologies potential danger and public health.
\indent Opinions become more precise with controversies. They made 39 arguments during the first session and 19 during the second one. Each argument was systematically discussed.
\indent The content of justifications leads to the following conclusions:
\begin{itemize}
 \item If a statement is tagged \textbf{proved}, the justification explains the statement.
\item If a statement is tagged \textbf{incorrect}, the justification is a negative reason.
 \item If a statement is tagged \textbf{with two criteria} ( proved and unclear or proved and incorrect), the justification shows a contradiction and suggests a new direction.
\item If a statement is tagged \textbf{not incorrect, unclear and not proved}, the justification is expression of a belief or of a
conviction.
\end{itemize}

\indent If these results are confirmed by more experiments, it means that we can guide users in justification of their choices of judgment. 41 annotations were well-formed ( i.e a judgment per level), the figure 14 presents the repartition of these annotations with respect to relations of sub-alternation of our structure.

\begin{figure}[H]
\begin{center}
 \includegraphics[scale=0.5]{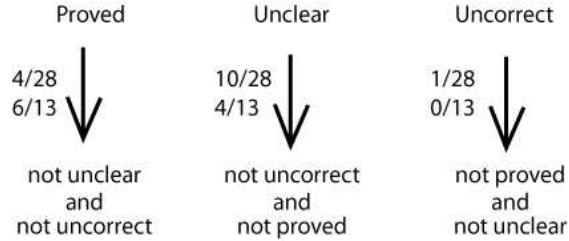}
\caption{\it Sub-alternation in Intermed}
\end{center}
\end{figure}

\subsection{Desire and pleasure}
\indent Notions of desire and pleasure have appeared during our experiments. Desire is an impulsion for users to participate. Pleasure is a satisfaction, a gratification. The third experiment was the most representative of the need to not neglect these notions. Every frustration (misunderstanding of the software, technical issues, lack of information) can block the participation but an ergonomic interface (nice and easy to use) can enhance desire to debate.
Pleasure depends on quality and goal of the debate. Reasoning and arguing, clarifying one’s own opinion and producing a summary of one’s argumentation are pleasurable activities. These two notions (desire and pleasure) deserve more attention in our future works. But we have seen we could mesure desire and pleasure with poll. Students of the Intermed experiment attributed an average mark of 4/5 for the quality of the debate and of 3,5/5 for the ergonomy and the interest of the website.

\section{Conclusion}
\indent Facing the lack of web 2.0 solutions to make debates of quality, we propose to create web tools founded on logic to enhance debates structure and comprehension. In our solution, arguments are justified judgments on statements. The experiments we presented show that our model of argumentation is natural and easy to adopt. The n-opposition theory is a geometrical tool we use to define relations between judgments and to categorize relations between arguments. That's a first step to describe reasonings. The third experiment showed that it is possible to define for each judgment an "illocutionary-act". We can associate an interpretation to each judgment. But, is it natural to represent a judgment with a logical modality? What kind of logic can express relations between judgments?
\\
\indent In our future experiments, we suggest the implementation of an help system that will suggest to users a form of justification for their judgments. We have to define notions of desire and pleasure in the context of the debate. We want to develop an agent founded on n-opposition and logic for helping users when reasoning.

\section{Acknowledgement}
\indent We thank JB.Soufron for its help during the first experiment. J.B.Soufron was responsible of the website betapolitique. Thank to V.Pinet and P.Thomaso and their students for testing our tools. Thank to N.Desquinabo and all the team of the Intermed project. Thank to Pikko-software society and specially its developers for technical help. Thank to J.Divol, N.Rodriguez, A.Gouaich for their help at LIRMM. And a special thank to M.Hornby for her help during translation of this paper.

\bibliographystyle{plain}
\bibliography{biblioJYB}
\end{document}